\documentclass[11pt,a4paper]{article}
\usepackage[hyperref]{emnlp-ijcnlp-2019}
\usepackage{times}
\usepackage{latexsym}
\usepackage{booktabs} 

\usepackage{url}
\newcommand{\cut}[1]{}

\usepackage{graphicx,xcolor} 
\usepackage{amsmath}
\usepackage{mathtools}
\usepackage{bm}
\usepackage{wrapfig,lipsum,booktabs}
\usepackage{amsfonts}

\newcommand{\deepspeare}{\textsc{Deep-speare}}
\newcommand{\deepspeares}{\textsc{Deep-speare's}}
\newcommand{\rhymem}{\textsc{Rhym-em}}
\newcommand{\method}{\textsc{Rhyme-gan}}

\newcommand{\methodlm}{\textsc{Rhyme-lm}}

\newcommand{\methodnostruct}{\textsc{Rhyme-gan-ns} }
\newcommand{\sonnetdata}{\textsc{Sonnet}}
\newcommand{\limerickdata}{\textsc{Limerick}}

\aclfinalcopy 

\title{Learning Rhyming Constraints using Structured Adversaries}

\author{Harsh Jhamtani$^1$, Sanket Vaibhav Mehta$^1$, Jaime Carbonell$^1$, Taylor Berg-Kirkpatrick$^2$ \\
$^1$ School of Computer Science, Carnegie Mellon University\\
$^2$ Department of Computer Science and Engineering, University of California San Diego \\
\tt{\{jharsh,svmehta,jgc\}@cs.cmu.edu, tberg@ucsd.eng.edu}
}

\date{}

\begin{document}

\maketitle
\begin{abstract}
Existing recurrent neural language models often fail to capture higher-level structure present in text: for example, rhyming patterns present in poetry. Much prior work on poetry generation uses manually defined constraints which are satisfied during decoding using either specialized decoding procedures or rejection sampling. The rhyming constraints themselves are typically not learned by the generator. We propose an alternate approach that uses a structured discriminator to learn a poetry generator that directly captures rhyming constraints in a generative adversarial setup. By causing the discriminator to compare poems based only on a learned similarity matrix of pairs of line ending words, the proposed approach is able to successfully learn rhyming patterns in two different English poetry datasets (Sonnet and Limerick) without explicitly being provided with any phonetic information.
\end{abstract}

\section{Introduction}

\begin{figure*}
    \centering
    \includegraphics[width=0.85\textwidth]{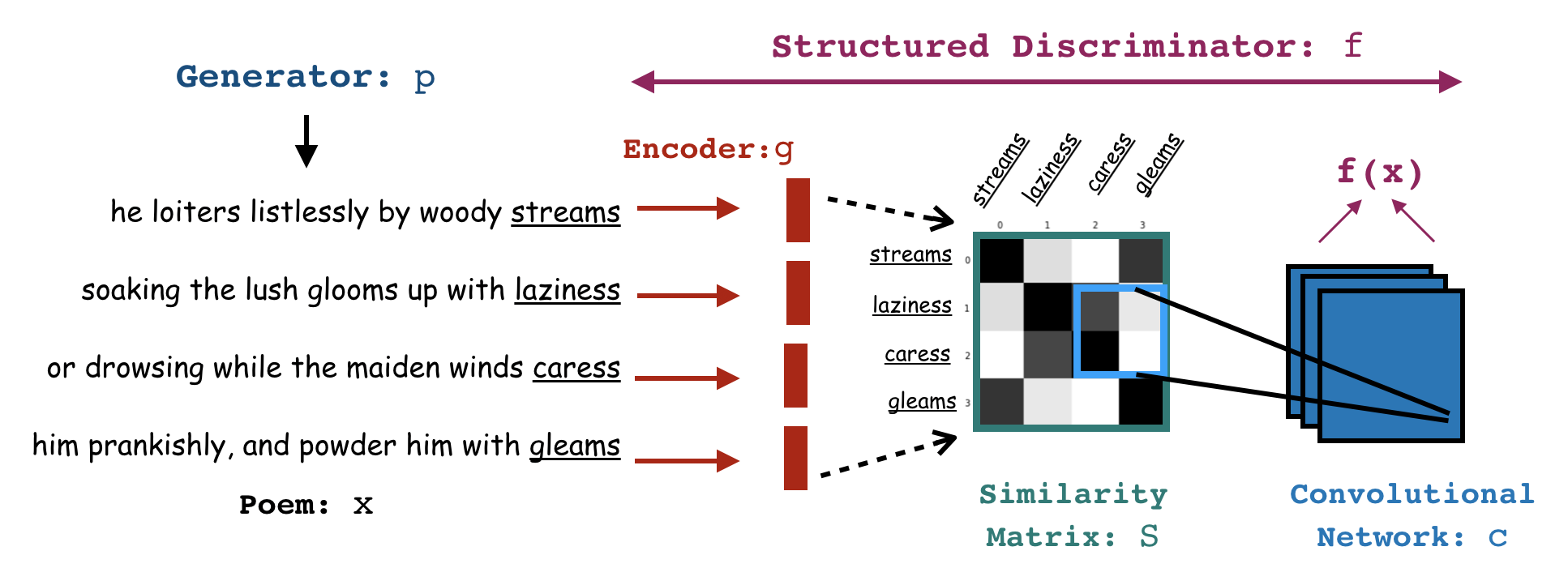}
    \caption{\small Model Overview: We propose a structured discriminator to learn a poetry generator in a generative adversarial setup. Similarities between pairs of end-of-line words are obtained by computing cosine similarity between their corresponding representations, produced by a learned character-level LSTM encoder.  The discriminator operates on the resulting matrix $S$ representing pair-wise similarities of end words. The proposed discriminator learns to identify rhyming word pairs as well as rhyming constraints present in the dataset without being provided phonetic information in advance. }
    \label{fig:model}
\end{figure*}

Many existing approaches to text generation rely on recurrent neural networks trained using likelihood on sequences of words or characters. However, such models often fail to capture overall structure and coherency in multi-sentence or long-form text \cite{bosselut2018discourse, holtzman2018learning}. To rectify this, prior work has proposed losses which encourage overall coherency or other desired behavior \cite{li2016deep,zhang2017sentence,bosselut2018discourse}. However, most of these approaches rely on manually provided definitions of what constitutes a good or suitable structure, thereby limiting their applicability.\cut{In this paper we explore neural text generation with simultaneously learning structural patterns in generated output. We choose to work with poetry generation since different forms of poetry often contain rich rhyming and rhythm pattern constraints.} In this paper we propose a method for English poetry generation that directly learns higher-level rhyming constraints as part of a generator without requiring strong manual intervention.  
Prior works on poetry generation have focused mostly on ad-hoc decoding procedures to generate reasonable poetry, often relying on pruning from a set of candidate outputs to encourage desired behavior such as presence of explicitly-defined rhyming patterns \cite{oliveira2017survey,ghazvininejad2018neural}.  

We propose an adversarial approach to poetry generation that, by adding structure and inductive bias into the discriminator, is able to learn rhyming constraints directly from data without prior knowledge. The role of the discriminator is to try to distinguish between generated and real poems during training. 
We propose to add inductive bias via the choice of discriminator architecture: We require the discriminator to reason about poems through pairwise comparisons between line ending words. These learned word comparisons form a similarity matrix for the poem within the discriminator's architecture. Finally, the discriminator evaluates the poem through a 2D convolutional classifier applied directly to this matrix. This final convolution is naturally biased to identify spatial patterns across word comparisons, which, in turn, biases learned word comparisons to pick up rhyming since rhymes are typically the most salient spatial patterns. \cut{We show in our experiments that without informing the model that the similarity function it should learn is about rhyming, the model induces an accurate rhyming metric and then learns natural rhyming patterns on top of the similarity matrix.}

Recent work by \newcite{lau2018deep} proposes a quatrain generation method that relies on specific domain knowledge about the dataset to train a classifier for learning the notion of rhyming: that a line ending word always rhymes with exactly one more ending word in the poem. This limits the applicability of their method to other forms of poetry with different rhyming patterns. They train the classifier along with a language model in a multi-task setup. Further, at generation time, they heavily rely on rejection sampling to produce quatrains which satisfy any valid rhyming pattern. 
In contrast, we find that generators trained using our structured adversary produce samples that satisfy rhyming constraints with much higher frequency.

Our main contributions are as follows: We introduce a novel structured discriminator to learn a poetry generation model in a generative adversarial setup.
We show that the discriminator induces an accurate rhyming metric and the generator learns natural rhyming patterns without being provided with phonetic information.
We successfully demonstrate the applicability of our proposed approach on two datasets with different structural rhyming constraints. 
Our poem generation model learned with the structured discriminator is more sampling efficient compared to prior work -- many fewer generation attempts are required in order to obtain an valid sample which obeys the rhyming constraints of the corresponding poetry dataset.  
\cut{The proposed discriminator learns the notion of rhyming and also learns to identify valid rhyming patterns from the data without being provided with phonetic information or specific domain knowledge about the dataset.}

\section{Method}

\cut{Poetry often involves different types of structural patterns and constraints \cite{oliveira2017survey,ghazvininejad2016generating,ghazvininejad2017hafez,lau2018deep}. }Many forms of poetry make use of rhyming patterns on line-ending words \cite{oliveira2017survey}. Therefore, to characterize a rhyming poem, a model needs (1) to know what it means to rhyme (2) to identify the specific permissible rhyming patterns for a particular poem type. For example, a limerick is a 5 line poem with a rhyming constraint of the type \textsc{AABBA}, i.e. the ends of the first, second, and fifth lines rhyme.
We consider an adversarial learning setup with a hierarchical language model and a structured discriminator, such that the discriminator is trained to distinguish between generated examples and training examples, and the generator is trained to \emph{fool} the discriminator. Our novel structured discriminator operates on a matrix which encodes a {\em learned} pair-wise similarity function of the line ending words. We refer to our model as \textbf{\method{}}. \\

\subsection{Neural Generation Model}
Our generator is a hierarchical neural language model (Figure \ref{fig:model}) that first generates a sequence of line-ending words, and thereafter generates the poem's lines conditioned on the ending words. We use recurrent neural networks for ending word generation as well line generation conditioned on ending words. Following prior work \cite{lau2018deep}, we generate words in each line in reverse order (i.e. right to left), and begin generation with the last line first. Let $\hat{x}$ represent a sample from the current generator distribution, denoted by $p_\theta$, where $\theta$ represents the generator parameters. 
We initialize the word embeddings in the generator with pre-trained word embeddings \cite{lau2018deep} trained on a separate non-sonnet corpus.\cut{ consisting of more than $34M$ words. }\\

\subsection{Structured Discriminator}
We introduce a structured discriminator, denoted by function $f_\phi(x)$, which outputs the probability that $x$ is a sample from the dataset as opposed to generated. Our architecture defines an intermediate matrix $S \in R^{T \times T}$, where $T$ denotes the number of lines in the poem, which encodes pair-wise similarities between line ending words in order to capture rhyming structure. The discriminator's output is determined by a two layer 2D convolutional neural network applied to $S$. Convolutional neural networks have been shown to capture local as well as global patterns in 2D data -- for example, images. Thus, our discriminator is composed of two main components: computation of a matrix $S$, and a convolutional neural network to classify the computed matrix $S$. The pair-wise computation provides a useful inductive bias to identify the notion of rhyming, whereas the convolutional network is a suitable choice to capture overall rhyming patterns. 

More specifically, let the words at the ends of lines in $x$ be denoted by $e$. The number of ending words will be same as the number of lines in $x$, which we denote as $T$.
We encode each ending word using a character-level LSTM \cite{hochreiter1997long} denoted by $g_{\phi_g}$, and use the last hidden state of the LSTM as a vector representation of the word. We let $S_{ij}$ be the cosine similarity between the representations of ending words $e_i$, $e_j$, given by following equation:
\begin{equation}
    S_{ij} = \frac{g(e_i) g(e_j) }{|g(e_i)| |g(e_j)|}
\end{equation}
The matrix $S$ is passed through a convolutional neural network composed with a linear layer, together denoted by $c_{\phi_c}$. The final output is passed through a sigmoid non-linearity, so that $f_\phi(x) \in [0,1]$. The value of $f_\phi(x)$ represents the discriminator's assessment of the probability that datum $x$ belongs to the \emph{real} dataset, rather than being a generated sample.
The discriminator's objective will train it to distinguish between a sample $x$ from training data $\mathcal{X}$, and a generated sample $\hat{x}$, in a binary classification setup. Specifically, we define the discriminator loss for $x,\hat{x}$ as follows:
\begin{equation}
     d(x,\hat{x};\phi) =  -\log(f_\phi(x)) -\log(1-f_\phi(\hat{x})) 
\end{equation}

\subsection{Learning}
Generator parameters $\theta$ and discriminator parameters $\phi$ are trained together under following objective:
{\small
\begin{equation}
     \min_\theta \left[ \mathbb{E}_{x \in \mathcal{X}} \left[ -\log p_\theta(x)   
     + \lambda \max_\phi \mathbb{E}_{ \hat{x} \sim p_\theta} [ -d(x,\hat{x}) ] \right] \right]
\end{equation}}
Note, in addition to using a traditional adversarial objective, we also include a likelihood term to help stabilize the generator. $\lambda$ is a hyperparameter which controls the relative weight of the two terms. Since sampling of $\hat{x}$ from generator involves discrete choices, we use the REINFORCE \cite{williams1992simple} algorithm to train the generator using learning signal from the adversarial loss term. The generator simultaneously gets an exact gradient from the likelihood portion of the objective.
We observe training is more stable when we pretrain the LSTM word encoder $g_{\phi_g}(.)$ part of the discriminator, along with a separate LSTM decoder, using an auto-encoding objective on words in the vocabulary. 

\section{Experiments and Results}

\subsection{Datasets}
We work with the Shakespeare \sonnetdata{} dataset \cite{lau2018deep} and a new \limerickdata{} corpus. 
Each sonnet in the Sonnet dataset is made up of 3 quatrains of 4 lines each, and a couplet. The dataset consists of 2685 sonnets in train, and 335 each in validation and test splits. 
The quatrains typically have one of the following rhyming structures: AABB, ABAB, ABBA, though some deviations are observed in the dataset. This may be because rhyming patterns are not always strictly followed in writing quatrains, and there are possible inaccuracies in the word pronunciation dictionaries used (e.g. some words can have multiple different pronunciations based on context).

A limerick  is a form of verse with five lines.\cut{, often humorous and sometimes rude} Limericks typically follow a rhyming pattern of AABBA. We collect limericks from an online collection\footnote{\url{http://hardsoft.us}. Accessed May 2019.}. Due to a large vocabulary in the full collection, we filter the dataset to retain only those limericks whose all the words are in a subset of 9K most frequent words. Our final dataset consists of $10,400$ limericks in train and $1300$ each in validation and test splits. 
We train and evaluate the models separately on each corpus.

\subsection{Poem Generator}

\begin{table}
    \centering
    \small
    \begin{tabular}{c|c c}
        Model & \multicolumn{2}{c}{Expected \#Samples} \\ 
         & \sonnetdata{} & \limerickdata{} \\ \hline
        \deepspeare{} &  $153.8$ & N/A \\
        \methodlm{} &  $169.5$ & $500$ \\
        \methodnostruct{} & $4.8$ & $26.6$ \\
        \textbf{\method{}} & $\mathbf{3.7}$ & $\mathbf{4.7}$
    \end{tabular}
    \caption{\small Sampling efficiency: We obtain 10K samples of poetry without additional intervention during decoding, and report the expected samples as inverse of the fraction of samples satisfying valid rhyming patterns for the corresponding dataset. Lower values are better.}
    \label{tab:sampling}
\end{table}

\noindent \textbf{Sampling efficiency}
We compute the expected number of samples needed before we sample a quatrain which satisfies one of the hand-defined rhyming patterns. 
Towards this end, we get 10K samples from each model without any constraints (except avoiding \emph{UNK} - unknown tokens). Following prior work \cite{lau2018deep}, words are sampled with a temperature value between $0.6$ and $0.8$.
We use CMU dictionary \cite{weide1998cmu} to look up the phonetic representation of a word, and extract the sequence of phonemes from the last stressed syllable onward. Two words are considered to be rhyming if their extracted sequences match \cite{parrish}. We consider a generated quatrain to have an acceptable pattern if the four ending words follow one of the three rhyming patterns: AABB, ABBA, ABAB. Similarly for \limerickdata{}, we consider only those samples to be acceptable which have line endings of the rhyming form AABBA.

We consider a baseline \textbf{\methodlm{}}, which has the same generator architecture as \method{} but is trained without the discriminator.
We also compare with \textbf{\methodnostruct{}} which uses a simpler non-structured discriminator. Specifically, it uses a discriminator which first runs a character-level encoder for each ending word - similar to \method{} - but then instead of computing pair-wise similarity matrix, it utilizes a LSTM on the sequence of the computed representations. 

As can be observed from  Table \ref{tab:sampling}, \method{} needs fewer samples than other methods to produce an acceptable quatrain or a limerick, indicating that it has learned natural rhyming structures more effectively from data.  
Note we do not report \deepspeare{} on Limerick due to their \sonnetdata{} specific assumption that for a given end-of-line word there is exactly one more rhyming word among other end-of-line words. Additionally, \methodnostruct{} performs worse than \method{}, and the difference in performance is more prominent in \limerickdata{}  -- demonstrating that the proposed structure in the discriminator provided useful inductive bias.  Note that compared to 4 line quatrains in \sonnetdata{}, \limerickdata{} has 5 line poems and has arguably more complex rhyming pattern constraints. \\

\noindent \textbf{Likelihood on held out data} We report negative log likelihood (NLL) on test splits (Table \ref{tab:likelihood}). For \sonnetdata{}, \method{} achieves a test set NLL of $3.98$. Our model without adversarial learning i.e. \methodlm{}, achieves a test set NLL of $3.97$. \deepspeare{} reports a test set NLL of $4.38$. Note that our language model is hierarchical while \deepspeare{} has a linear model.\cut{ For \limerickdata{}, NLL on test split is $3.49$ for \method{} and $3.48$ for \methodlm{}.} The NLL for \methodlm{} and \method{} are very similar, though \method{} gets much better sampling efficiency scores than \methodlm{}. \\

\begin{table}
    \centering
    \small
    \begin{tabular}{c|c c}
        Model & \multicolumn{2}{c}{NLL} \\ 
         & \sonnetdata{} & \limerickdata{} \\ \hline
        \deepspeare{} & $4.38$ & N/A \\
        \textbf{\methodlm{}} &  $\mathbf{3.97}$ & $\mathbf{3.48}$ \\
        \method{} & $3.98$ & $3.49$ \\ 
    \end{tabular}
    \caption{\small Held out negative log likelihood per token for poems in test split.}
    \label{tab:likelihood}
\end{table}

Our generator implementation is largely based on that of \citet{lau2018deep}. The main difference is that we first generate all the line-ending words and then condition on them to generate the remaining words. The change was made to make it more amenable to our proposed discriminator. However, our hierarchical language model (\methodlm) performs worse than \deepspeare{} as per sampling efficiency. Therefore, structured discriminator is the driving factor behind the observed improvement with \method. However, committing to the ending words of all lines before completing preceding lines can be a limitation, and addressing it is a possible future direction.

\subsection{Analyzing Learned Discriminator}
We probe the the word representations $g(.)$ to check if rhyming words are close-by in the learned manifold. We consider all pairs of words among the ending words in a quatrain/limerick, and label each pair to be rhyming or non-rhyming based on previously stated definition of rhyming.\cut{We skip the words not available in the CMU pronunciation dictionary.}
If the cosine similarity score between the representations of pairs of words is above a certain threshold, we predict that word pair as rhyming, else it is predicted as non-rhyming. We report F1 scores for the binary classification setup of predicting word-pairs to be rhyming or not. 
We consider some additional baselines: \textbf{\rhymem{}} \cite{reddy2011unsupervised} uses latent variables to model rhyming schemes, and train parameters using EM. \textbf{\textsc{grapheme-k}} baselines predict a word pair as rhyming only if the last $K=\{1,2,3\}$ characters of the two words are same. 

For \sonnetdata{} data, we observe that \method{} obtains a F1 score of $0.90$ (Table \ref{tab:rhyming}) on the test split (threshold chosen to maximize f1 on dev split). We repeat the above analysis on the \limerickdata{} dataset and observe an F1 of $0.92$ for \method{}. 
\deepspeare{} model \cut{\cite{lau2018deep}} reports F1 of $0.91$ on \sonnetdata{}. As stated earlier, \deepspeares{} model is not amenable to \limerickdata{} - we do compare though with the max-margin classifier in \deepspeare{} model trained on \limerickdata{} which gets F1 score of $0.81$. The scores are understandably lower since the AABBA pattern in limericks is not amenable to \sonnetdata{} specific assumptions made in \deepspeare{} model. 
\cut{Note that \deepspeare{} incorporates dataset specific rhyming pattern information in it's training objective.} On the other hand, \method{} achieves high F1 scores for both the datasets without incorporating any domain specific rhyming pattern information. 

\method{} performs much better than \rhymem{} and \textsc{grapheme-k} baselines. \rhymem{} does not perform well - probably because it operates at word-level and fails to generalize.  Note that \textbf{\methodnostruct{}} gets F1 score of $0.85$ in case of \sonnetdata{} dataset and $0.87$ for \limerickdata{}. These values are lower than corresponding scores for \method{}, demonstrating that the proposed structure in the discriminator was useful in learning the notion of rhyming.

\begin{table}
    \centering
    \footnotesize
    \begin{tabular}{c|c|c}
       Model & \sonnetdata{} & \limerickdata{} \\ \hline
        \textsc{grapheme-1} & $0.71$ & $0.79$ \\
        \textsc{grapheme-2} & $0.78$ & $0.79$  \\ 
        \textsc{grapheme-3} & $0.69$ & $0.67$ \\ 
        \rhymem{} & $0.71$ & $0.73$  \\  
        \deepspeare{}/\textsc{Max-Margin} &  $0.91$ & $0.81$ \\
        \methodnostruct{} & $0.85$ & $0.87$ \\
        \textbf{\method{}} & $\mathbf{0.90}$ & $\mathbf{0.92}$ 
    \end{tabular}
    \caption{\small Rhyming probe: We use the cosine similarity score of the learned representations to predict a word pair as rhyming or not, and report F1 score for this classification task. \rhymem{} \cut{\cite{reddy2011unsupervised}} is an unsupervised rhyming pattern discovery method. \textsc{grapheme-k} baselines predict based on exact match of last $k$ characters. }
    \label{tab:rhyming}
\end{table}

\subsection{Human Evaluations}
Following prior work \cite{lau2018deep}, we requested human annotators to identify the human-written poem when presented with two samples at a time - a quatrain from the Sonnet corpus and a machine-generated quatrain, and report the annotator accuracy on this task. Note that a lower accuracy value is favorable as it signifies higher quality of machine-generated samples. Using 150 valid\cut{/accepted} samples (i.e. samples belonging to one of the allowed rhyming patterns), we observe $56.0\%$ annotator accuracy for \method{}, and $53.3\%$ for \deepspeare{} -- indicating that the post-rejection sampling outputs from the two methods are of comparable quality (the difference in annotator accuracy is not statistically significant as per McNemar's test under $p<0.05$).  If we use \cut{unfiltered (}pre-rejection samples, we observe $60.0\%$ annotator accuracy for \method{}, and $81.3\%$ for \deepspeare{} (the difference being statistically significant as per McNemar's test under $p<0.05$) -- indicating that unfiltered samples from \method{} are of higher quality compared to \deepspeare{}.

\section{Related Work}

Early works on poetry generation mostly used rule based methods \cite{gervas2000wasp,wu2009new,oliveira2017survey}. More recently, neural models for poetry generation have been proposed \cite{zhang2014chinese,ghazvininejad2016generating,ghazvininejad2017hafez,hopkins2017automatically,lau2018deep,liu2019rhetorically}.
\newcite{yan2013poet} retrieve high ranking sentences for a given user query, and repeatedly swap words to satisfy poetry constraints.
\newcite{ghazvininejad2018neural} worked on poetry translation using an unconstrained machine translation model and separately learned Finite State Automata for enforcing rhythm and rhyme. 
Similar to rhyming and rhythm patterns in poetry, certain types of musical compositions showcase rhythm and repetition patterns, and some prior works model such patterns in music generation \cite{walder2018neural,jhamtanirepeatgan2019}. 
Generative adversarial learning \cite{goodfellow2014generative} for text generation has been used in prior works \cite{fedus2018maskgan, wang2018no, wang2019deep, rao2019answer}, though has not been explored with regard to the similarity structure proposed in this paper.

\section{Conclusions}
In this paper we have proposed a novel structured discriminator to learn a poem generator. The generator learned utilizing the structured adversary is able to identify rhyming structure patterns present in data, as demonstrated through the improved sampling efficiency. Through the rhyming classification probe, we demonstrate that the proposed discriminator is better at learning the notion of rhyming compared to baselines. 

\section*{Acknowledgements}
We are thankful to anonymous EMNLP conference reviewers for providing valuable feedback. We thank Graham Neubig for useful discussions on poetry generation. This project is funded in part by the NSF under grant 1618044 and by the NEH under grant HAA-256044-17.

\bibliography{emnlp2019}
\bibliographystyle{acl_natbib}

\section*{Appendix: Additional Implementation Details}
We use PyTorch framework to implement the models\footnote{Trained model weights and output samples will be released at \url{github.com/harsh19/Structured-Adversary}}.
We use 100 dimensional word embeddings and 128 dimensional LSTM hidden size.
We experiment with multiple values of $\lambda$ and found that $\lambda=0.1$ worked well for both datasets. Recall that $\lambda$ is the weight of the discriminator term in the overall loss as depicted in equation 3.
As mentioned earlier, we separately pre-train the word encoder part of discriminator using an auto-encoder setup i.e. words are encoded by the encoder, and a separate LSTM is used to predict the input sequence of characters from the last hidden state of the encoder. We observe that pre-training provided small gains in sampling efficiency in case of \limerickdata{} dataset.

\end{document}